\documentclass{article}
\usepackage{spconf,amsmath,graphicx, bm, tabularx}
\usepackage{subcaption} 
\usepackage{multirow}
\usepackage{adjustbox}
\usepackage{balance}
\usepackage{xcolor}
\usepackage{hyperref}

\title{On the exploitation of DCT-traces in the Generative-AI domain}
\name{Orazio Pontorno, Luca Guarnera, Sebastiano Battiato}
\address{Department of Mathematics and Computer Science, University of Catania, Italy}
\begin{document}
%
\maketitle
\begin{abstract}

Deepfakes represent one of the toughest challenges in the world of Cybersecurity and Digital Forensics, especially considering the high-quality results obtained with recent generative AI-based solutions. Almost all generative models leave unique traces in synthetic data that, if analyzed and identified in detail, can be exploited to improve the generalization limitations of existing deepfake detectors. In this paper we analyzed deepfake images in the frequency domain generated by both GAN and Diffusion Model engines, examining in detail the underlying statistical distribution of Discrete Cosine Transform (DCT) coefficients. Recognizing that not all coefficients contribute equally to image detection, we hypothesize the existence of a unique ``discriminative fingerprint", embedded in specific combinations of coefficients. To identify them, Machine Learning classifiers were trained on various combinations of coefficients. In addition, the Explainable AI (XAI) LIME algorithm was used to search for intrinsic discriminative combinations of coefficients. Finally, we performed a robustness test to analyze the persistence of traces by applying JPEG compression. The experimental results reveal the existence of traces left by the generative models that are more discriminative and persistent at JPEG attacks. Code and dataset are available at \href{https://github.com/opontorno/dcts_analysis_deepfakes}{\texttt{github/opontorno/dcts\_analysis\_deepfakes}}.
\end{abstract}
\begin{keywords}
Synthetic Traces, Deepfakes, Multimedia Forensics
\end{keywords}
\vspace{-0.3cm}

\section{Introduction}
\vspace{-0.2cm}
\label{sec:intro}
In the age of Artificial Intelligence, the phenomenon of deepfakes has emerged as a significant challenge. These sophisticated digital manipulations, created through Generative Adversarial Networks (GANs)~\cite{goodfellow2014generative} and Diffusion Models (DMs)~\cite{ho2020denoising}, have raised concerns about their ability to generate deceptively realistic images. Distinguish real from synthetic images has become a priority in the field of Multimedia Forensics~\cite{piva2013overview,battiato2016multimedia}, with crucial implications for security, information and privacy. In the current scientific landscape, the scientific community has predominantly focused on developing sophisticated detection solutions. Several studies~\cite{guarnera2020preliminary,marra2019gans} demonstrated that generative models leave unique traces on the produced content. 
Exploiting the presence of these anomalies in multimedia content, different approaches in the literature (mainly based on deep learning techniques) are able to excellently define not only the nature of the data under analysis~\cite{wang2020cnn} (real or deepfake), but also establish the specific model used for creation~\cite{guarnera2022exploitation}.  The main limitations of these approaches concern the possibility of generalization in any context (multimodal data, different semantics, etc.). A more analytical analysis of the traces would lead to the creation of ``targeted" detectors capable of overcoming these limitations. In this paper, we propose an analytical method to study and establish patterns in the frequency domain to be defined as unique traces of the synthetic data.
The research investigates the potential of AC statistics coefficients derived from the Discrete Cosine Transform (DCT) to discern among real images, those produced by DMs, and those created by GAN technologies. It states that specific combinations of these features contain unique discriminative traces for each category of images, suggesting not all coefficients equally contribute to the classification process. 
A preliminary analysis focused on manually selecting specific subsets of coefficients, used to train standard machine learning classifiers, including K-Nearest Neighbours (K-NN), Random Forest and Gradient Boosting. 
We also exploited the Explainable AI (XAI) algorithm LIME~\cite{ribeiro2016should} to determine the most significative combinations of coefficients using a deep learning approach, allowing us to investigate the distinctive features in a targeted and technologically advanced manner. Finally, once the most discriminative traces were identified, they were tested for persistence to JPEG attacks.

\noindent The main contributions of this paper are:
\begin{itemize}
   \vspace{-0.2cm}
   \item One of the first statistical analyses conducted on such a large collection of images generated by both GANs and DMs.
   \vspace{-0.3cm}
   \item A preliminary exploration to determine a differentiating trace between the three classes (real, GANs, DMs) in the DCT domain.
\end{itemize}
\vspace{-0.2cm}
\noindent The paper is structured as follows. An overview of the main Deepfakes Detection methods based on intrinsic trace identification is reported in Section~\ref{sec:rel_works}. Section~\ref{sec:dataset} gives a detailed description of dataset. Section~\ref{sec:prop_meth} describes the proposed approaches, while the experimental results are detailed in Section~\ref{sec:exp_res}. 
Finally, Section~\ref{sec:conclusion} concludes the paper with some hints for future investigation in this domain.
\vspace{-0.3cm}

\section{Related Works}
\vspace{-0.2cm}
\label{sec:rel_works}

Most deepfake detection algorithms are based on the capture of intrinsic traces to detect them. 
Approaches working in the spatial domain have been shown to be able to capture intrinsic traces that can egregiously discriminate real images from deepfakes. Specifically, using the Expectation-Maximization algorithm, Guarnera et al.~\cite{guarnera2020fighting} demonstrated that the correlation between pixels in deepfake images turns out to be different than in real images: the convolutional layers fail to reproduce the acquisition pipeline of common devices (image creation is different). Therefore, such a trace (pixel correlation) not only turns out to be different than in real images, but also allows discrimination between different generative architectures (since the GAN architectures themselves are different). In the same domain, McCloskey et al.~\cite{mccloskey2019detecting} identified another discriminant trace defined by the statistics of color channel curves. The latter, in most of the images created by generative models in the literature, turn out to be ``undersampled" compared to real images acquired by the various acquisition devices. Recent studies have highlighted the effectiveness of techniques based on the frequency domain for identifying abnormal traces left by generative models during the creation step, achieving notable results. Particularly noteworthy are deepfake detectors that employ the DCT. These methods include the application of DCT directly to images, as illustrated by Joel et al.~\cite{JOEL_ICML_2020}, and the use of features extracted from DCT blocks (following the JPEG compression pipeline), as demonstrated in~\cite{concas2022tensor}. 
Giudice et al.~\cite{Giudice_JI_2021} proposed another approach based on DCT to establish the so-called \textit{GAN-specific frequencies} (GSFs), in order to identify for each involved architecture the most discriminating and characterizing frequency for the specific generative model. The $\beta$ statistics inferred from the distribution of AC coefficients~\cite{lam2000mathematical} were key to recognizing the data generated by the GAN engine.
\begin{figure}[t!]
    \centering
\includegraphics[width=0.4\textwidth]{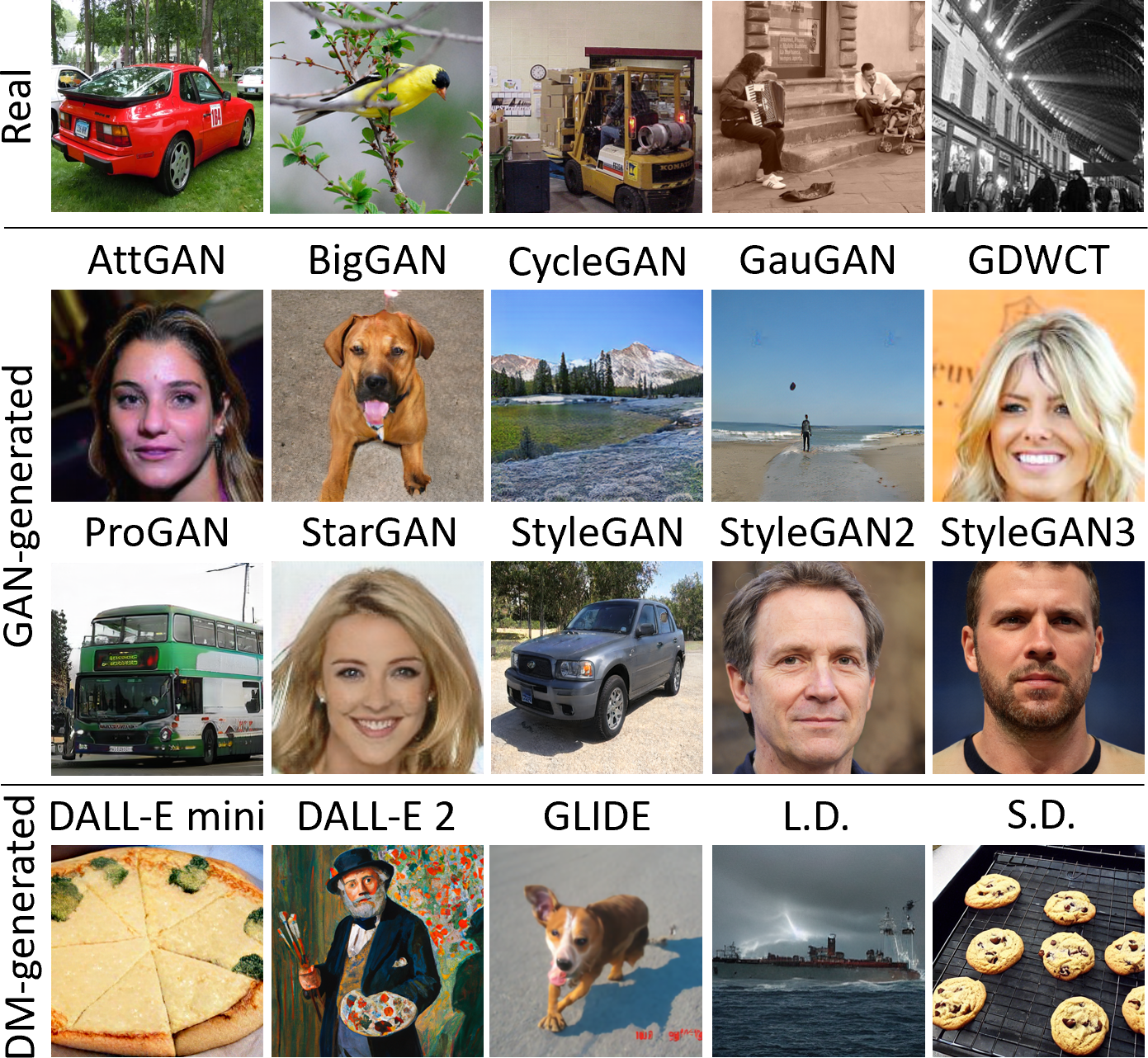}
\vspace{-0.2cm}
    \caption{Sample images of the various involved datasets.}
    \vspace{-0.5cm}
    \label{fig:db}
\end{figure}
The analysis in the Fourier domain~\cite{zhang2019detecting} is the base of most methods in literature~\cite{guarnera2020preliminary,marra2019gans} in order to discover and study those discriminative traces on synthetic data with respect to real content. 
Other approaches~\cite{Durall_CVPR_2020,Dzanic_NEURIPS_2020} based on the analysis of images in the frequency domain demonstrated that it is possible to identify the specific anomalous fingerprints, achieving competitive results in detecting deepfakes. Both approaches, namely spatial-based and frequency-based, have shown considerable efficacy in detecting and characterizing the underlying digital fingerprint of the employed generative model.

\vspace{-0.2cm}

\section{Dataset Details}
\vspace{-0.2cm}
\label{sec:dataset}

\begin{table}[t!]
    \centering
    \begin{footnotesize}
    \begin{adjustbox}{max width=0.4\textwidth}
    \begin{tabular}{cccc}
        \hline
        \textbf{Type} & \textbf{\begin{tabular}[c]{@{}c@{}}Architecture \\ Name\end{tabular}} & \textbf{\begin{tabular}[c]{@{}c@{}}\# Image \\ Used\end{tabular}} & \multicolumn{1}{c}{\textbf{Source}} \\ 
        \hline
        \multirow{9}{*}{\begin{tabular}[c]{@{}c@{}}GAN \\ GENERATED\end{tabular}} & GauGAN~\cite{park2019gaugan} & 4000 & $\star$\\
        \cline{2-4}
         & BigGAN~\cite{brock2018large} & 2600 & \tiny{\cite{corvi2023detection}}\\
        \cline{2-4}
         & ProGAN~\cite{karras2017progressive} & 1000 & \tiny{\cite{guarnera2020fighting}; \cite{corvi2023detection}} \\
        \cline{2-4}
         & StarGAN~\cite{choi2018stargan} & 6848 & \tiny{\cite{guarnera2020fighting}; \cite{corvi2023detection}}\\
        \cline{2-4}
         & AttGAN~\cite{he2019attgan} & 6005 & \tiny{\cite{Giudice_JI_2021}}\\ 
        \cline{2-4}
         & GDWCT~\cite{cho2019image} & 3367 & \tiny{\cite{guarnera2020fighting}}\\ 
        \cline{2-4}
         & CycleGAN~\cite{zhu2017unpaired} & 1047 & \tiny{\cite{guarnera2020fighting}}\\
        \cline{2-4}
         & StyleGAN~\cite{karras2019style} & 4705 & \tiny{\cite{Giudice_JI_2021}; \cite{guarnera2020fighting}; \cite{corvi2023detection}}\\ 
        \cline{2-4}
         & StyleGAN2~\cite{karras2020analyzing} & 7000 & \tiny{\cite{guarnera2020fighting}; ~\cite{corvi2023detection}}\\ 
        \cline{2-4}
         & StyleGAN3~\cite{karras2021alias} & 1000 & \tiny{\cite{corvi2023detection}}\\ 
        \hline
        \multirow{5}{*}{\begin{tabular}[c]{@{}c@{}}DM \\ GENERATED\end{tabular}} & DALL-E 2~\cite{ramesh2022hierarchical} & 3423 & $\star$; \tiny{\cite{leotta2023not}} \\ 
        \cline{2-4}
         & DALL-E MINI & 1000 & \tiny{\cite{corvi2023detection}}\\  
        \cline{2-4}
         & Glide~\cite{nichol2021glide} & 2000 & $\star$; \tiny{\cite{corvi2023detection}} \\ 
        \cline{2-4}
         & Latent Diffusion~\cite{rombach2022high} & 4000 & $\star$; \tiny{\cite{corvi2023detection}} \\ 
        \cline{2-4}
         & Stable Diffusion & 5000 & $\star$; \tiny{\cite{corvi2023detection}; \cite{cocchi2023unveiling}} \\ 
        \hline
         REAL  & - & 19341 & $\star$ \\
        \hline
    \end{tabular}
    \end{adjustbox}
    \end{footnotesize}
    \vspace{-0.2cm}
    \caption{Number of images used per category (Real, GAN, DM) in the experimental phase. The fourth column presents the sources from which the images were collected. 
    The $\star$ indicates the images collected or generated by us.}
    \vspace{-0.3cm}
    \label{tab:dataset}
\end{table}

In the assembly of the dataset used for conducting the experiments, a comprehensive and diverse approach was employed. The dataset consists of a total of $72.334$ images, categorized as follows: $19.334$ real images~\footnote{The real images were collected in parts from CelebA~\cite{liu2015faceattributes}, FFHQ~\cite{karras2019style} and the articles cited in the Table~\ref{tab:dataset}.}, $37.572$ images generated by GAN architectures and $15.423$ images produced by DMs (Fig.~\ref{fig:db} shows some examples of the involved dataset). The dataset was initially divided into training and testing sets following a proportion of $85\%$ and $15\%$, respectively. There is a quantitative bias in favor of GAN-generated images, resulting in a class imbalance. To mitigate potential overfitting problems due to this disparity, the Under-Sampling strategy was implemented during the training phase. This technique aims to balance the classes, ensuring that the machine learning model does not develop a preference for the numerically preponderant class. Aligned with the research objectives, the heterogeneity of the dataset was prioritized. As a result, the images were selected to minimize the significance of their semantic content, focusing instead solely on the underlying generative architecture. This approach ensures that the dataset is representative of various image generation techniques, irrespective of their specific content. This choice aims to emphasize variety in the dataset's composition, with images sourced from multiple origins, each characterized by different tasks and image creation methodologies. Table~\ref{tab:dataset} provides a detailed analysis of the dataset~\footnote{Note that Stable Diffusion and DALL-E MINI were taken from the links \href{https://github.com/CompVis/stable-diffusion}{github.com/CompVis/stable-diffusion} and \href{https://github.com/borisdayma/dalle-mini.}{github.com/borisdayma/dalle-mini}}, presenting both the total number of images for each architectural category and the diverse sources from which they were procured.

\vspace{-0.3cm}

\section{Proposed Method}
\vspace{-0.2cm}
\label{sec:prop_meth}
The proposed methodology consists of training some standard machine learning algorithms, implementing different combinations of the intrinsic AC coefficients statistics in order to find the most discriminative one. 


\vspace{-0.4cm}
\subsection{Extraction and Analysis of AC statistic}
In the initial phase, the coefficients resulting from the application of the DCT (Equation~\ref{eq:DCT}) were calculated for each digital image under analysis. This process involves the preliminary partition of each image into non-overlapped blocks of size $8\times8$ pixels, with the DCT transform subsequently applied to each block.
\begin{equation}
    \label{eq:DCT}
    \small
    DCT(u, v) = \frac{C(u) \cdot C(v)}{4} \sum_{x=0}^{7} \sum_{y=0}^{7} f(x, y) \cdot t(x,u) \cdot t(y,v)
\end{equation}
where $C(p)=\begin{cases} \frac{1}{\sqrt{2}} & \text{if } p=0 \\ 1 & \text{otherwise} \end{cases}$, $t(z, \epsilon) = \cos\left[\frac{(2z+1)\epsilon\pi}{16}\right]$ and $f(x,y)$ is the pixel intensity at coordinates $(x, y)$ in the image block. These coefficients are then sorted into an $8\times8$ matrix in the classic `zig-zag' order, according to the standard JPEG compression pipeline.

\begin{figure}
    \centering
    \includegraphics[width=.4\textwidth]{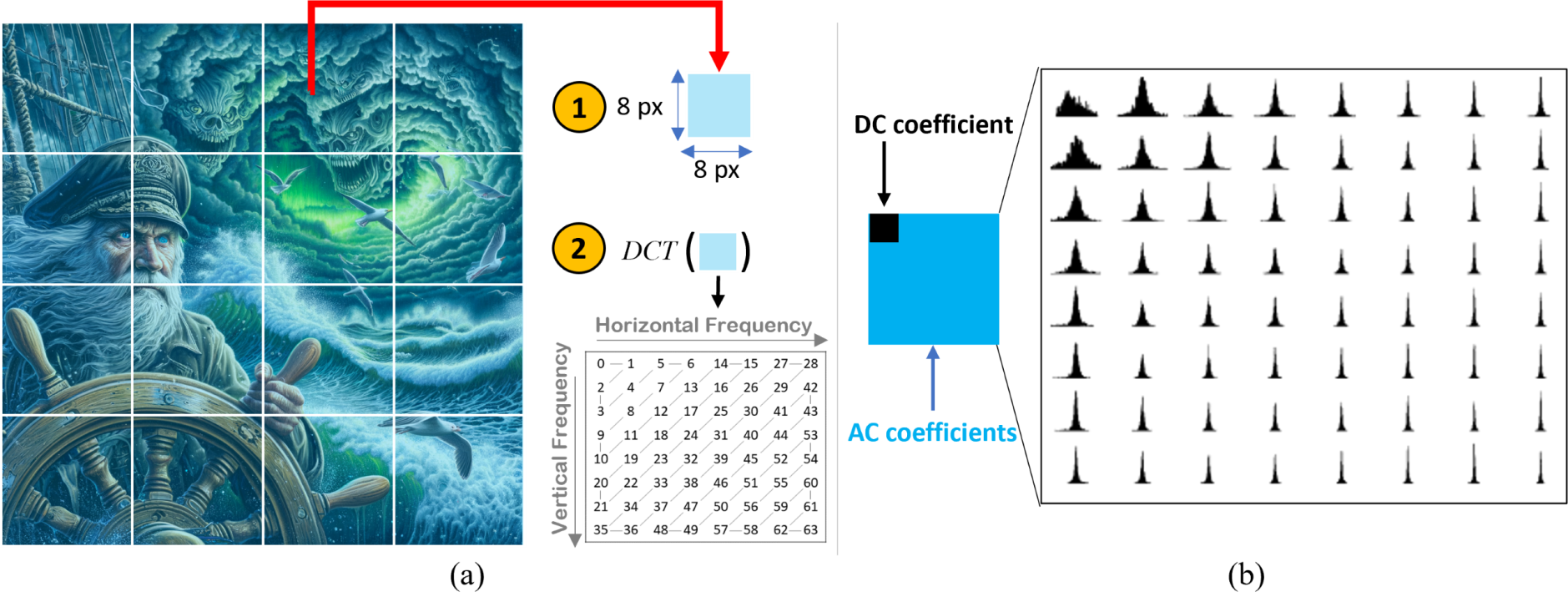}\vspace{-0.3cm}
    \caption{
    (a) DCT applied on non-overlapping $8\times8$ blocks of the input image. (b) Sampling of the overall Gaussian and Laplacian distributions from each block (Image source:~\cite{lam2000mathematical}).}
    \vspace{-.3cm}
    \label{fig:dct_sortering}
\end{figure}

The values of the DCT coefficients can be regarded as the relative amount of the two-dimensional spatial frequencies contained in the 64 input points (Fig.~\ref{fig:dct_sortering}(a)). The first of these coefficients (top left, position $(0,0)$) is 
\textit{DC} and roughly represents the average brightness level of the entire image. All other coefficients are called \textit{AC}, representing the specific bands of the image itself. The 63 AC coefficients, which vary in frequency and amplitude, providing a detailed picture of the brightness variations within the block. Lam L. Y. et al.~\cite{lam2000mathematical} demonstrated that the \textit{DC} coefficient can be modelled with a Gaussian distribution while the \textit{AC} coefficients can be modelled effectively by means of a parametric Laplacian distribution (Fig.~\ref{fig:dct_sortering}(b)), which PDF is described by Equation~\ref{eq:laplacian}, characterized by the location parameter $\mu$ and scale parameter $\beta^{AC}$. The $\beta^{AC}$ value depends on the specific context of the data embedding information useful for different task~\cite{farinella2015representing}, $\beta^{AC}$ is commonly estimated as $\sigma / \sqrt{2}$, where $\sigma$ corresponds to the standard deviation of the AC coefficient distributions. Finally we have:
\vspace{-0.1cm}
\begin{equation} \label{eq:laplacian}
    P(x|\mu, \beta) = \frac{1}{2\beta} \exp\left(-\frac{|x - \mu|}{\beta}\right)
\end{equation}

\begin{figure}
    \centering
    \includegraphics[width=0.3\textwidth]{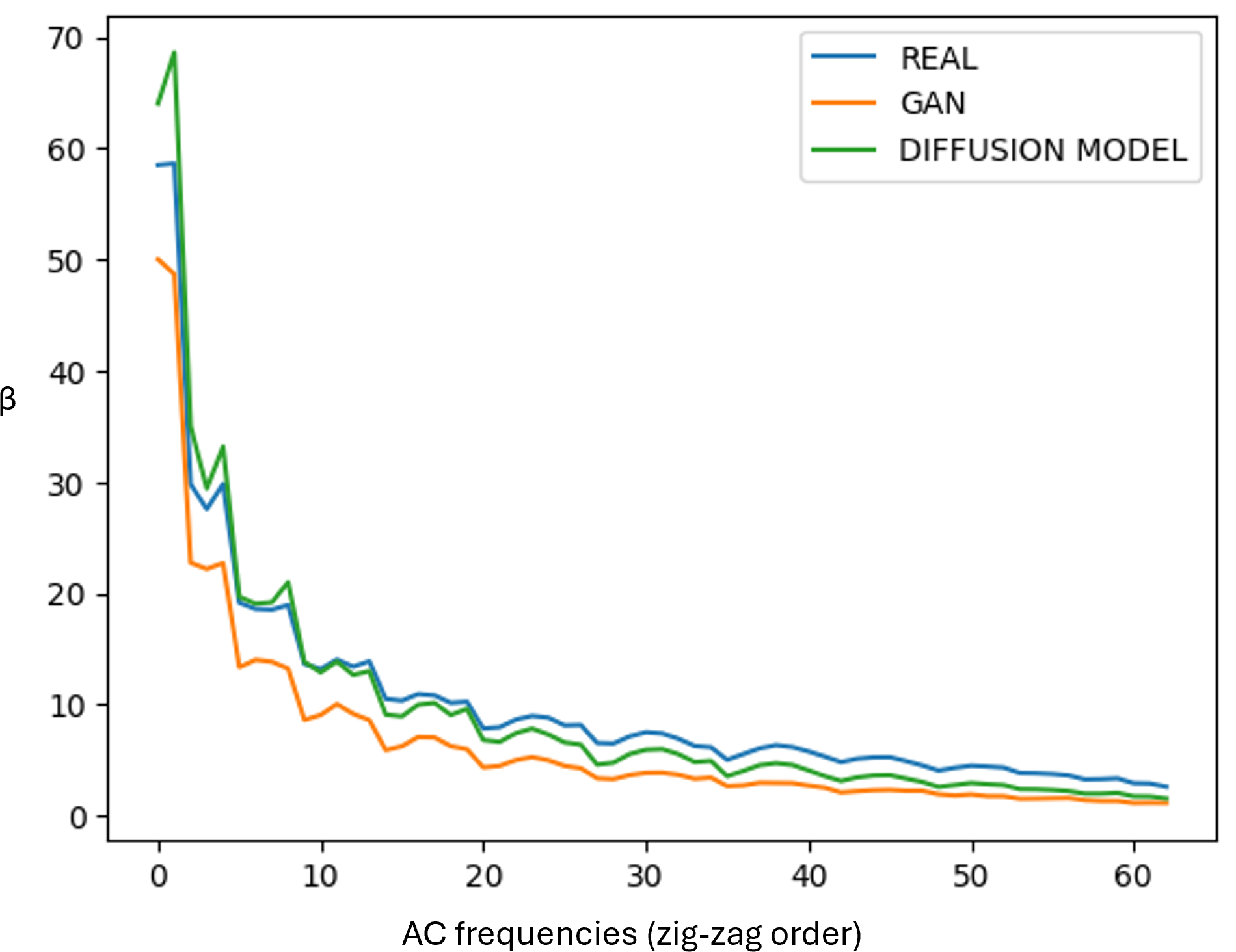}
    \vspace{-0.3cm}
    \caption{Average distributions of $\beta^{AC}$ for each class. AC coefficients represented according to the classical ``zig-zag" ordering.} 
    \vspace{-0.4cm}
\label{fig:ave_beta_distr}
\end{figure}

The proposed method focuses specifically on $\beta^{AC}$ coefficients, which showed significant discriminating properties to distinguish real images from those generated through GAN engines~\cite{Giudice_JI_2021}. Once $\beta^{AC}$ coefficients had been extracted, we analyzed their average trend for each involved technology (Fig.~\ref{fig:ave_beta_distr}), classifying the coefficients according to the specific categories of the images to which they belong, in order to better understand their distinguishing characteristics. 
From Fig.~\ref{fig:ave_beta_distr}, there is a clear distinction between the three classes in terms of the magnitude of their $\beta^{AC}$ coefficients. This suggests that $\beta^{AC}$ coefficients, taken together or as a subset, can be used to differentiate between real images, GAN, and DMs. On the left-hand side of the graph, corresponding to low frequencies, we notice a greater variance than in the high frequencies. In this area, the overall trend of these coefficients extracted from images generated by DMs is almost completely superimposed on that of the real images, thus suggesting a difficult distinction between these two classes. As we shift our attention to the right, towards the coefficients representing the high frequencies, we notice both a convergence for all three classes and above all a separation between the class of real images and those generated by DMs. This ``tiered'' structure could be exploited by Machine Learning algorithms to distinguish involved classes.

\vspace{-0.25cm}
\subsection{Machine learning approach with manual selection} \label{par:mac_hands}
The present research focuses on the identification of the most discriminating subset of $\beta^{AC}$ coefficients. To this end, three classical machine learning classifiers were selected: K-Nearest Neighbors (K-NN), Gradient Boosting, and Random Forest. K-NN was chosen for its ability to identify patterns based on the proximity of elements, assuming that images produced by the same architecture share similar coefficients. Gradient Boosting and Random Forest, both classified as ensemble methods, were selected for their robustness to variations in input data and their structure based on decision trees. The latter allows the previously analyzed layered structure to be exploited through the mechanism of partitioning the input dataset.
Initially, training was conducted by considering the full range of 63 $\beta^{AC}$ coefficients, establishing a base value of accuracy. Subsequently, the process was iterated using only a subset of the mentioned coefficients. 
The primary objective was to assess whether the accuracy maintained with a reduced subset of the coefficients was comparable to that obtained with the entire set. 
From the analysis of the distribution of $\beta^{AC}$ coefficients described in the previous section, an iterative procedure was adopted to identify the optimal subset of $\beta^{AC}$ coefficients. Initially, the first $k$-coefficients were selected, with $k$ varying from 2 to 30. At a later stage, the focus shifted to subsets consisting of the last $t$-coefficients, with $t$ decreasing from 35 to 2. Finally, we focused on the central part of the distribution, where the mean values of the coefficients of the images generated by the Diffusion Model clearly deviate from those of the real images. At this stage, subsets of the $\beta^{AC}$ considered were extracted by taking increasing interval range from the 31st coefficient: $[31-z,31+z]$ with $z=1,2,\dots,15$ It is relevant to note that for the training of each algorithm, ``Random Search'', a technique used for parameter selection in machine learning models by randomly selecting combinations of parameters from a given search space, was implemented in combination with ``Cross Validation'', using 3 folds.

\vspace{-0.3cm}
\subsection{Machine learning approach with LIME selection} \label{par:mac_lime}
The research on identifying the most discriminative subset of $\beta^{AC}$ coefficients, as described previously, involved the manual selection of potential subsets. 
However, it acknowledges the possibility that the most relevant subset might not necessarily consist of consecutive $\beta^{AC}$ coefficients.
In order to select the best $\beta^{AC}$-subsets, we considered the use XAI algorithm LIME~\cite{ribeiro2016should} to discern a more discriminative subset of coefficients. The 63 $\beta^{AC}$ were given to a customized neural network composed by three hidden layers with $256, 128, 64$ neurons respectively, each followed by the ReLU activation function, in order to classify them with respect to the three classes: real, GAN-generated and DM-generated. 

The initial phase exploited the NN's ability to identify non-linear relationships among the $\beta^{AC}$ coefficients for accurate classification of the generating architecture. Subsequently, the LIME algorithm was applied for each corrected classification to determine the contribution of each coefficient to the prediction. The average contribution $C^{avg}_{\beta^{AC}_i}$ of each $\beta^{AC}$ was calculated by averaging individual contributions across all correctly classified test images (Equation~\ref{eq:avg_c_lime}).
\vspace{-0.2cm}
\begin{equation} \label{eq:avg_c_lime}
    C^{avg}_{\beta^{AC}_i} = \frac{1}{N}\sum_{j=1}^{N}c_{\beta^{AC}_i}(p_j) \quad i=1,\dots,63
    \vspace{-0.2cm}
\end{equation}

\begin{figure}[t!]
    \centering
\includegraphics[width=0.3\textwidth]{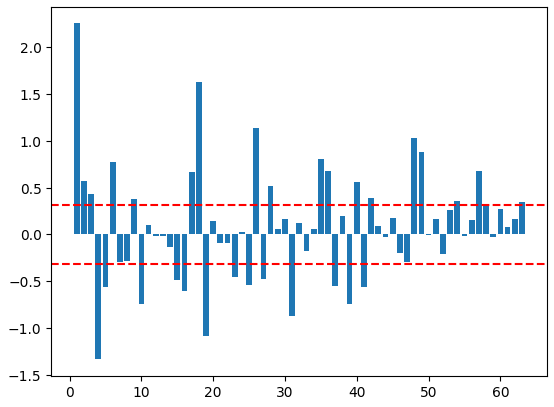}
\vspace{-0.3cm}
    \caption{Average contributions $C^{avg}_{\beta^{AC}_i}$ computed using LIME. Dotted lines indicate the median $\underset{j}{Med}(|C^{avg}_{\beta^{AC}_j}|)$.}
    \vspace{-0.5cm}
    \label{fig:lime_all}
\end{figure}

where $p_j$ is the $j$-th correct prediction made by the network, $N$ is the total number of correct predictions and $c_{\beta^{AC}_i}(p_j)$ is the contribution of $\beta_i$ to the prediction $p_j$, as calculated by LIME.
Fig.~\ref{fig:lime_all} illustrates the average contributions $C^{avg}_{\beta^{AC}_i}$ of coefficients in each successful classification. Each value indicates the contribution (positive or negative) that the coefficient had in the correct prediction. It can be observed that some coefficients had a greater absolute contribution than others. Based on this observation, we identified two distinct subsets of discriminative $\beta^{AC}$ coefficients. The first subset, called \textit{POS-LIME} (Equation~\ref{eq:pos_lime}), comprises coefficients with a positive average contribution, while the second one, called \textit{ABS-LIME} (``ABsolutely Significative'', Equation~\ref{eq:abs_sign}) includes those whose average contribution's absolute value exceeds the median $\underset{j}{Med}(|C^{avg}_{\beta^{AC}_j}|)$ of all average contributions. It's important to note that we opted for the median over the mean as a reference point due to the mean's susceptibility to skewness from contributions near zero. In detail we have:
\vspace{-0.2cm}
\begin{equation} \label{eq:pos_lime}
\vspace{-0.3cm}
    \text{\small{POS-LIME}} = \{ \beta^{AC}_i : C^{avg}_{\beta^{AC}_i} > 0 \} 
\end{equation}
\begin{equation} 
\label{eq:abs_sign}
\small
    \text{\small{ABS-LIME}} = \{ \beta^{AC}_i : |C^{avg}_{\beta^{AC}_i}| > \underset{j}{Med}(|C^{avg}_{\beta^{AC}_j}|) \} 
    \vspace{-0.2cm}
\end{equation}

The chosen subsets were employed for the advanced training of the three machine learning algorithms.

\begin{table*}[t!]
    \centering
    \begin{adjustbox}{max width=1\textwidth}
    \renewcommand{\arraystretch}{1.3}
    \begin{tabular}{|c |ccccc| ccccc |ccccc|}
        \hline
        & \multicolumn{5}{c|}{\uppercase{\large{K-NN}}}&\multicolumn{5}{c|}{\uppercase{\large{Random Forest}}}&\multicolumn{5}{c|}{\uppercase{\large{Gradient Boosting}}} \\
        \textbf{Acc/F1 (\bm{$\%$})} & & \multicolumn{4}{c|}{\textit{JPEG Compression Test}} & & \multicolumn{4}{c|}{\textit{JPEG Compression Test}} & & \multicolumn{4}{c|}{\textit{JPEG Compression Test}} \\ \cline{3-6} \cline{8-11} \cline{13-16}
         & \textbf{RAW} & \textbf{QF90} & \textbf{QF70} & \textbf{QF50} & \textbf{QF30} & \textbf{RAW} & \textbf{QF90} & \textbf{QF70} & \textbf{QF50} & \textbf{QF30} & \textbf{RAW} & \textbf{QF90} & \textbf{QF70} & \textbf{QF50} & \textbf{QF30} \\
        \hline
        $\beta^{AC}_{ALL}$ & 68.76/66.05 & 68.45/65.65 & 64.34/61.53 & 56.46/53.55 & 47.78/45.47 & 79.59/77.81 & 41.75/39.58 & 33.30/30.23 & 29.54/26.81 & 29.44/26.37 & 82.57/80.81 & 44.68/44.08 & 35.10/32.84 & 32.22/29.26 & 30.58/26.93 \\
        \hline
        \hline
        $\beta^{AC}_{1:28}$ & 64.56/61.87 & 64.61/62.02 & 63.82/61.22 & 63.16/60.13 & 61.43/58.60 & 71.13/69.11 & 69.32/67.42 & 56.34/55.64 & 43.63/43.44 & 34.58/33.94 & 72.33/70.32 & 70.86/68.83 & 57.63/57.09 & 43.71/44.07 & 35.01/34.58 \\
        \hline
        $\beta^{AC}_{1:29}$ & 64.96/62.30 & 64.86/62.31 & 63.82/61.26 & 63.22/60.25 & 61.17/58.31 & 71.18/69.18 & 69.49/67.54 & 54.91/54.27 & 41.19/40.80 & 32.84/31.85 & 73.03/70.99 & 71.38/69.30& 56.56/56.26 & 41.49/41.69 & 33.35/31.91 \\
        \hline
        $\beta^{AC}_{1:30}$ & 64.80/62.16 & 64.66/62.12 & 64.08/61.44 & 63.17/60.14 & 60.35/57.39 & 71.78/69.84 & 69.67/67.81 & 54.11/53.55 & 40.33/39.97 & 32.39/31.31 & 73.29/71.25 & 71.84/69.82 & 56.23/56.00 & 41.37/41.75 & 33.02/31.41 \\
        \hline
        \hline
        $\beta^{AC}_{16:46}$ & 69.62/67.08 & 67.70/65.41 & 48.09/48.22 & 32.25/31.78 & 24.78/23.57 & 74.10/72.12 & 53.42/53.20 & 34.35/31.93 & 28.54/25.19 & 29.32/22.98 & 76.71/74.66 & 53.31/53.47 & 36.21/33.96 & 31.10/26.90 & 29.68/22.54 \\
        \hline
        $\beta^{AC}_{17:45}$ & 70.21/67.55 & 67.86/65.50 & 47.13/47.40 & 31.84/31.30 & 24.00/22.75 & 74.44/72.45 & 50.71/50.58 & 34.28/31.74 & 29.52/25.50 & 29.15/21.32 & 76.49/74.39 & 53.89/54.01 & 36.12/33.79 & 31.32/27.12 & 29.69/22.58 \\
        \hline
        $\beta^{AC}_{18:44}$ & 69.65/66.76 & 66.51/64.10 & 45.23/45.55 & 30.42/29.69 & 23.74/22.25 & 73.93/71.85 & 51.46/51.36 & 33.98/31.52 & 29.22/25.49 & 28.83/21.22 & 76.06/74.00 & 54.79/54.84 & 35.51/33.04 & 31.41/27.50 & 29.66/23.16 \\
        \hline
        \hline
        $\beta^{AC}_{29:63}$ & 72.91/70.04 & 54.56/54.41 & 31.63/30.68 & 26.55/24.94 & 23.51/21.94 & 80.10/78.19 & 40.34/37.98 & 34.28/31.19 & 31.73/28.47 & 29.89/25.98 & 81.73/79.79 & 42.34/41.02 & 35.01/32.18 & 32.65/29.33 & 30.26/26.02 \\
        \hline
        $\beta^{AC}_{30:63}$ & 73.08/70.15 & 54.06/53.90 & 31.36/30.23 & 26.14/24.44 & 23.54/21.81 & 80.01/78.08 & 40.44/38.08 & 34.53/31.43 & 31.86/28.63 & 29.90/26.05 & 81.54/79.60 & 42.21/40.93 & 35.09/32.28 & 32.31/28.89 & 30.38/26.27 \\
        \hline
        $\beta^{AC}_{31:63}$ & 73.45/70.56 & 53.70/53.53 & 30.77/29.52 & 26.09/24.43 & 23.63/21.78 & 79.73/77.78 & 40.40/38.03 & 34.23/31.10 & 32.02/28.79 & 29.91/26.28 & 81.48/79.57 & 42.25/40.90 & 34.77/31.89 & 32.53/29.12 & 30.17/25.96 \\
        \hline
        \hline
        $\beta^{AC}_{POS-LIME}$ & 67.74/64.77 & 67.23/64.22 & 61.77/59.00 & 53.24/50.81 & 44.86/42.87 & 80.05/78.16 & 41.92/39.85 & 33.42/30.25 & 30.23/27.49 & 29.02/26.06 & 82.21/80.34 & 44.62/43.84 & 34.72/32.00 & 31.48/28.62 & 30.64/27.03\\
        \hline
        $\beta^{AC}_{ABS-LIME}$ & 68.07/65.25 & 67.51/64.64 & 64.81/61.91 & 59.89/56.94 & 53.05/50.30 & 78.08/76.03 & 43.54/41.97 & 32.98/30.04 & 29.79/27.14 & 29.46/26.33 & 80.68/78.68 & 45.33/44.45 & 34.49/32.12 & 31.42/29.06 & 30.58/27.10\\
        \hline
    \end{tabular}
    \end{adjustbox}
    \vspace{-0.3cm}
    \caption{Accuracy (\%) and F1 Score (\%) values obtained during the testing phase. 
    Each row reports the experimental results performed for each specific range of $\beta^{AC}$ coefficients. 
    }
    \vspace{-0.3cm}
    \label{tab:results}
\end{table*}

\begin{figure*}[t!]
    \centering
    \begin{subfigure}[b]{0.32\textwidth} 
        \centering
        \includegraphics[width=\textwidth]{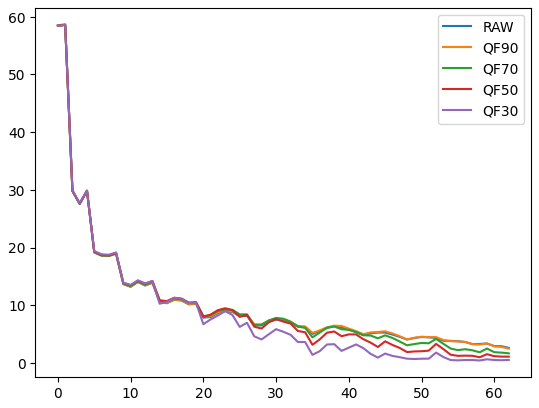}
        \caption{}
        \label{fig:figure1}
    \end{subfigure}
    \hfill 
    \begin{subfigure}[b]{0.32\textwidth}
        \centering
        \includegraphics[width=\textwidth]{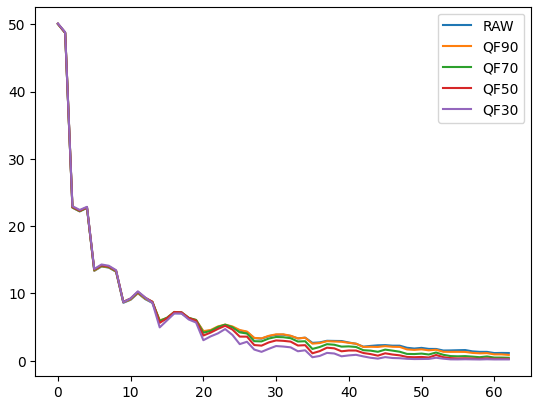}
        \caption{}
        \label{fig:figure2}
    \end{subfigure}
    \hfill
    \begin{subfigure}[b]{0.32\textwidth}
        \centering
        \includegraphics[width=\textwidth]{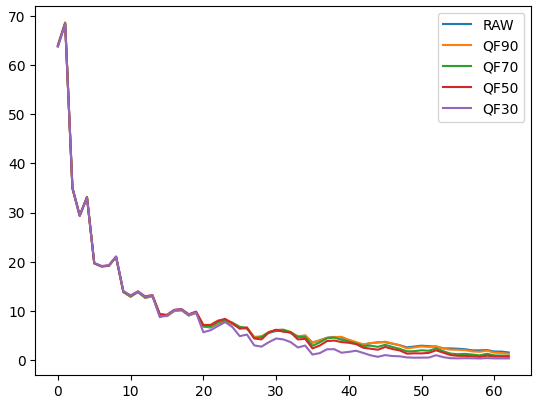}
        \caption{}
        \label{fig:figure3}
    \end{subfigure}
    \vspace{-.4cm}
    \caption{The three graphs show how the average distribution of $\beta^{AC}$ coefficients changes as the QF decreases, for real (a), GAN (b), and DM images (c). 
    The high-frequency components are gradually nullified as the QF increases during compression.}
    \vspace{-.3cm}
    \label{fig:changing_qf}
\end{figure*}
\vspace{-0.2cm}

\section{Experimental Results}
\vspace{-0.2cm}
\label{sec:exp_res}
Our study extends from the analysis of the existence of such traces, contained in subsets of $\beta^{AC}$ coefficients, in RAW images to the study of their persistence after JPEG compression. The experiments conducted are therefore divided into two phases: in the first phase, the three machine learning models are trained using different subsets of $\beta^{AC}$ coefficients extracted from RAW images. These models are then tested to infer the nature images using only a proper combination of $\beta^{AC}$: if the extracted metrics approach or exceed those obtained from using the full range of coefficients, then we can infer a high discriminative degree of this subset. The second step consists in analyzing the persistence at increasingly intense JPEG compressions of the discriminative degree of these subsets.

\vspace{-0.4cm}
\subsection{Research of most discriminative $\beta^{AC}$ in RAW images}
Following the process described in Section~\ref{par:mac_lime}, the two subsets were calculated using LIME: 
\vspace{-.1cm}
\[ \text{POS-LIME} = \{ \beta^{AC}_{i} \quad \forall i \in \text{idx}_\text{POS-LIME} \} \]
\vspace{-.6cm}
\[ \text{ABS-LIME} = \{ \beta^{AC}_{i} \quad \forall i \in \text{idx}_\text{ABS-LIME} \} \]
where $\text{idx}_\text{POS-LIME}$ and $\text{idx}_\text{ABS-LIME}$ are the sets of coefficient indices belonging to the \text{POS-LIME} and \text{ABS-LIME} sets respectively (See Fig.~\ref{fig:lime_all}). 
In the ``RAW" columns of Table~\ref{tab:results}, are reported the values obtained from the testing phase of the models trained with the different subsets of $\beta^{AC}$, taken as described in Sections~\ref{par:mac_hands} and~\ref{par:mac_lime}. An analysis of the results obtained shows that the best results have been obtained by ensemble classifiers using the subsets of coefficients dictated by LIME. Table~\ref{tab:results} clearly shows that, despite subset $\beta^{AC}_{POS-LIME}$ having half the coefficients, the models' accuracy when trained on this subset is comparable to or greater than that achieved with all coefficients. In general, it is possible to notice how in RAW images the last coefficients, corresponding to the high frequencies of the image, seem to have a discriminative degree higher than the others, thus strengthening our initial thesis that the ``tiered structure" that assumes the distribution of $\beta^{AC}$ can be exploited by classifiers based on decision trees in order to discern the nature of the image.

\vspace{-0.4cm}
\subsection{Analysis trace persistence at JPEG attacks}
\vspace{-0.1cm}
The second analysis begins with the compression of test images to different Quality Factors (QF): 90, 70, 50, and 30. Fig.~\ref{fig:changing_qf} shows how the distribution of $\beta^{AC}$ changes as QF decreases. It is easy to observe that the stronger the compression, the more the high frequency is gradually reduced. Experiments to analyze the persistence of the discriminative degree of coefficient subsets are shown in Table~\ref{tab:results} under the heading ``\textit{JPEG Compression Test}". These reveal how effectively this reduction of high frequencies is also reflected in the general performance of the trained models with subsets that contained the last coefficients. Ensemble methods, based on decision trees, can no longer discern the nature of digital images: in other words, the discriminatory trace vanishes. On the other hand, the performance of the K-NN model, although starting from an initial value of low accuracy, is more resistant to compression attacks. Moreover, despite the fact that in compressed images the first coefficients prove to be more discriminative, the K-NN manages to keep most of the discriminatory trace, contained in the subset ABS-LIME, even during compressions of increasingly intense images. Therefore, although all coefficients $\beta^{AC}$ can identify well the nature of synthetic data created by different generative technologies, a detailed analysis of only the low frequencies $\beta^{AC}_{1:28}$, through the use of XAI methods, could lead to identifying more robust unique traces capable of categorizing a specific generative model.

\vspace{-0.2cm}

\section{Conclusion and Future Works}
\vspace{-0.2cm}
\label{sec:conclusion}
In this paper, we carried out a preliminary analysis of the $\beta^{AC}$ coefficients extrapolated from the Laplacian distributions sampled from the DCT coefficients to discover intrinsic traces in synthetic content. Experimental results demonstrated that in RAW digital images this trace is contained largely in the high frequencies of the image, captured by the last $\beta^{AC}$ coefficients. However, this tends to fade as soon as the image is compressed. The use of the LIME algorithm allowed us to deepen the search by suggesting two different subsets of coefficients, taken scattered across the entire range of these, which have been shown to have both a high degree of discrimination in RAW images and good persistence, suggesting that more advanced distance-based models can indeed make better use of these coefficients. The preliminary analysis just conducted has shown how low frequencies manage to retain their discriminative degree even under the attacks of JPEG compression, a standard compression used in many social networks. It is therefore in the low frequencies that we will further investigate using XAI techniques to search for that trace that is able to be as discriminative as possible among the three classes (real, GAN generated, and DM generated) and at the same time persistent to compression.
\vspace{-0.3cm}

\section{Acknowledgment}
\vspace{-0.2cm}
\label{aknowledgement}
Orazio Pontorno is a PhD candidate enrolled in the National PhD in Artificial Intelligence, XXXIX cycle, organized by Università Campus Bio-Medico di Roma. Acknowledge financial support from: PNRR MUR project PE0000013-FAIR. This research is supported by Azione IV.4 - ``Dottorati e contratti di ricerca su  tematiche dell’innovazione" del nuovo Asse IV del PON Ricerca e Innovazione 2014-2020 “Istruzione e ricerca  per il recupero - REACT-EU”- CUP: E65F21002580005.

\bibliographystyle{IEEEbib.bst}
\vspace{-0.2cm}
\balance{
\bibliography{sample.bib}
}
\end{document}